\documentclass[conference]{IEEEtran}
\usepackage{cite}
\usepackage{amsmath,amssymb,amsfonts}
\usepackage{caption}  
\usepackage{graphicx}
\usepackage{textcomp}
\usepackage{makecell} 

\usepackage{hyperref}

\usepackage{xcolor}
\usepackage[utf8]{inputenc}
\usepackage{algorithm}
\usepackage{algpseudocode}
\def\BibTeX{{\rm B\kern-.05em{\sc i\kern-.025em b}\kern-.08em
    T\kern-.1667em\lower.7ex\hbox{E}\kern-.125emX}}
\begin{document}
\title{Multi-Robot Distributed Optimization for Exploration and Mapping of Unknown Environments using Bioinspired Tactile-Sensor  }
\author{\IEEEauthorblockN{Roman Ibrahimov}
\IEEEauthorblockA{\textit{Department of Mechanical Engineering} \\
\textit{University of California at Berkeley}\\
Berkeley, CA, US \\
roman.ibrahimov@berkeley.edu}
\and
\IEEEauthorblockN{Jannik Matthias Heinen}
\IEEEauthorblockA{\textit{Department of Mechanical Engineering} \\
\textit{University of California at Berkeley}\\
Berkeley, CA, US \\
jannik\_heinen@berkeley.edu }
}
\maketitle
\begin{abstract}
This project proposes a bioinspired multi-robot system using Distributed Optimization for efficient exploration and mapping of unknown environments. Each robot explores its environment and creates a map, which is afterwards put together to form a global 2D map of the environment. Inspired by wall-following behaviors, each robot autonomously explores its neighborhood, based on a tactile sensor, similar to the antenna of a cockroach, mounted on the surface of the robot. Instead of avoiding obstacles, robots log collision points when they touch obstacles. This decentralized control strategy ensures effective task allocation and efficient exploration of unknown terrains, with applications in search-and-rescue, industrial inspection, and environmental monitoring. The approach was validated through experiments using e-puck robots in a simulated 1.5 × 1.5 m environment with three obstacles. The results demonstrated the system's effectiveness in achieving high coverage, minimizing collisions, and constructing accurate 2D maps [\href{https://youtu.be/tZ4up4lfhaM}{\textit{a link for video descroption}}]. 
\end{abstract}

\vspace{-3mm}
\section{Introduction}

\vspace{-2mm}

Exploration and mapping of unknown environments is a critical task in various fields, including search-and-rescue operations, industrial inspection, and environmental monitoring \cite{chatziparaschis2020aerial, dunbabin2012robots}. Autonomous multi-robot systems have emerged as a promising solution, offering scalability, redundancy, and efficiency over single-robot approaches. Effective exploration, however, requires innovative sensing mechanisms and control strategies to navigate complex, obstacle-laden environments. Bioinspiration has proven to be a valuable approach in robotics, with nature providing elegant and robust solutions to challenging problems \cite{lei2022bio}. Among these, the tactile sensory capabilities of American cockroaches, particularly their ability to sense and respond to physical contact using antennae, offer a compelling model for robotic navigation in confined and cluttered spaces \cite{cowan2006task}.

This paper proposes a bioinspired multi-robot system that leverages tactile sensing and distributed optimization for efficient exploration and mapping of unknown environments. Robots equipped with tactile sensors, inspired by the antenna of cockroaches, navigate by logging collision points with obstacles rather than avoiding them. These logged points are used to create local maps, which are afterwards set together to form a global map. This approach was implemented on a group of e-puck robots in the Webots simulation environment, where the robots successfully built a 2D map of the environment, accurately capturing the layout of obstacles.
\begin{figure}[t]
    \centering
    \includegraphics[width=\linewidth]{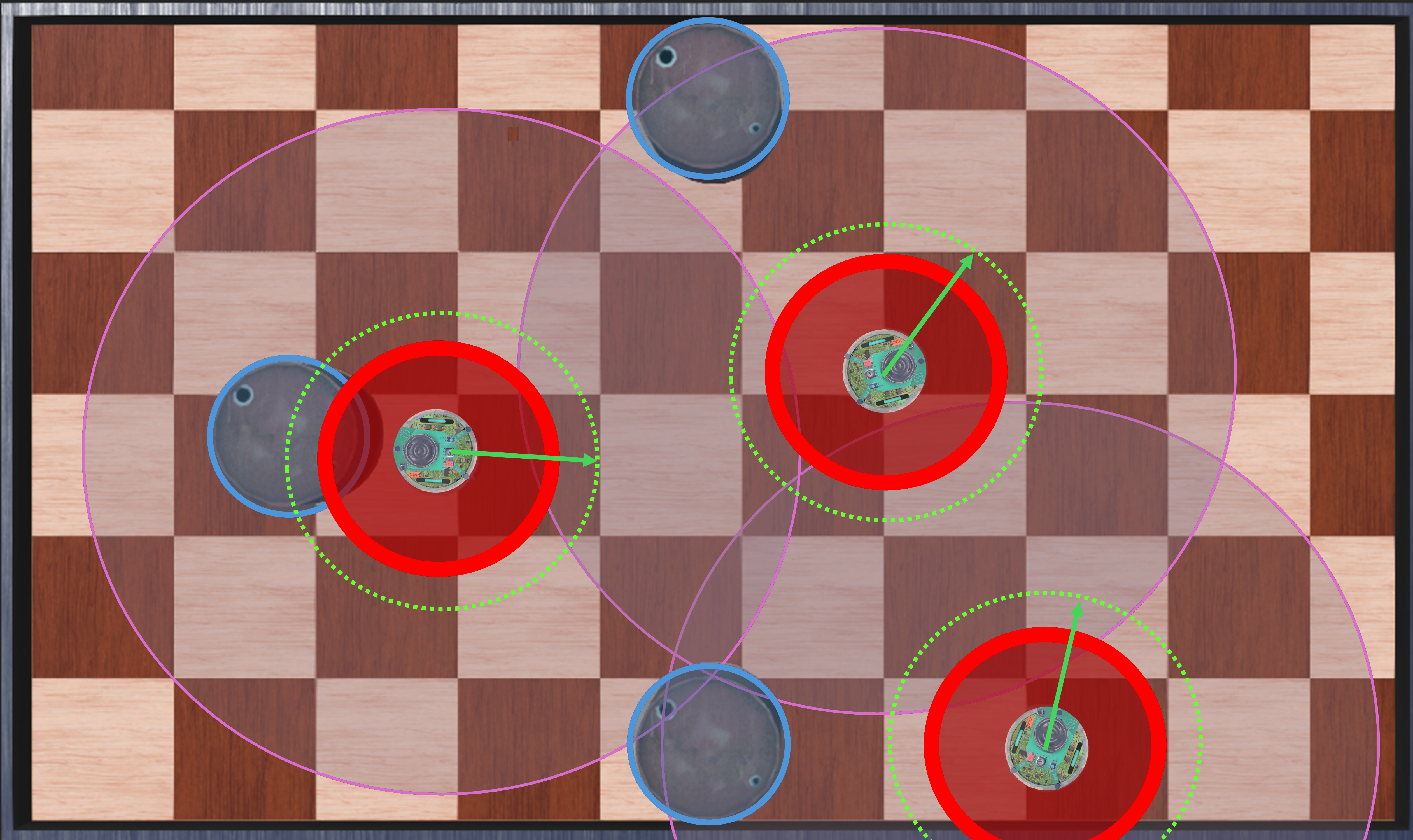}
    \caption{Multi-Robot Exploration of an Unknown Environment: The magenta circle represents each robot's communication radius, the red circle shows its sensing range for obstacle detection, and the green circles are candidate goal points. The green arrow indicates the selected trajectory. Robots detecting blue obstacles move backward and choose a new goal.}\label{Fig:FlightSequence}
\end{figure}



\section{Related Work}
A big majority of the current state-of-the-art research to address the mapping problem considers the environment as a set of opaque objects that can be sensed by optical sensors. Such sensors that are used in this research are mainly cameras and LiDAR sensors \cite{mulgaonkar2020tiercel}.  For example, in \cite{ebadi2020lamp}, the researchers equipped two autonomous underground Husky A200 rovers with LiDAR sensors to explore the network of underground tunnels. In similar paper \cite{scaramuzza2014vision}, a swarm of quadcopters are equipped with camera sensors to build a 3D map of the firefighters' training center. However, in the environments with transparent surfaces or in the scenarios with rain, dust, smoke or fog, the quality of the map with the above-mentioned sensors drops considerably. In \cite{goodin2019predicting}, researchers found out that the range of the LiDAR sensor considerably drops when the intensity of the rain increases. LiDAR sensors in a vast majority of mapping applications ignore glass or transparent surfaces and as a result the robot considers the space free-movable space \cite{foster2013visagge}. In a likewise scenario \cite{yamada2019vision}, it was shown that the quality of the image is lost in the rainy weather conditions despite the rain removal techniques. In addition, camera-based mapping algorithms often struggle with rapidly changing lighting conditions, which can lead to feature scarcity and loss of tracking \cite{camera_reflection}. A common approach to mitigate the limitations of standalone sensor solutions is to fuse sensor information, relying on the most reliable data from each sensor in specific scenarios. For example, combining camera and LiDAR sensors \cite{camera_lidar, camera_lidar2} or integrating LiDAR and IMU data \cite{lidar_imu} are popular strategies. The camera-LiDAR combination is particularly effective, as the camera captures detailed environmental textures while the LiDAR ensures mapping robustness against weather fluctuations and varying lighting conditions. Meanwhile, IMUs paired with either LiDAR or camera sensors can bridge short-term tracking interruptions, though they are less suitable for extended periods due to significant error accumulation. However, a key limitation of sensor fusion in practice is that it tends to be more costly than single-sensor solutions and challenging to calibrate \cite{yarovoi2024review}.


\section{Problem Formulation}

The task is to enable a group of \( N \) robots to autonomously explore and map an unknown 2D environment \( \mathcal{W} \subset \mathbb{R}^2 \) while detecting obstacles and collaboratively constructing a global map.

\vspace{-2mm}
\subsection{Environment Representation}
\vspace{-2mm}
The environment \( \mathcal{W} \subset \mathbb{R}^2 \) represents a continuous 2D space in which the robots operate. The workspace consists of two primary components: obstacles and free space. Obstacles are unknown regions that the robots aim to detect and log during their exploration, while the free space represents traversable areas that the robots can navigate. These components define the nature of the exploration task and the mapping objective.

The unknown obstacles in the environment are represented as a set of discrete points
$
\mathcal{O} = \{ \mathbf{o}_1, \mathbf{o}_2, \dots, \mathbf{o}_M \},
$
where \( \mathbf{o}_j \in \mathcal{W} \) denotes the position of an obstacle point. The total number of obstacle points is given by \( M \), which depends on the complexity of the environment. Identifying these obstacle points is a key goal of the exploration process.

Conversely, the free space in \( \mathcal{W} \) represents the regions that robots can safely navigate and explore. As the robots traverse \( \mathcal{W} \), they classify regions as either obstacle or free, progressively reducing the unexplored areas in the environment.

The exploration process results in the construction of a global map \( \mathcal{M} \) of the environment. This map associates every point \( \mathbf{x} \in \mathcal{W} \) with a specific state, which indicates whether the point is an obstacle, part of the free space, or remains unexplored $
\mathcal{M} = \{ (\mathbf{x}, \text{state}(\mathbf{x})) \mid \mathbf{x} \in \mathcal{W} \}.
$
Here, the function \( \text{state}(\mathbf{x}) \) assigns one of three possible classifications to each point \( \mathbf{x} \) such that $
\text{state}(\mathbf{x}) \in \{\text{Obstacle}, \text{Free}, \text{Unexplored}\}.
$
Initially, the state of most points in \( \mathcal{W} \) is classified as "Unexplored." As the robots navigate through the environment and encounter obstacles or traverse free space, the state of the corresponding points is updated to either "Obstacle" or "Free," depending on their observations. The objective of the exploration task is to reduce the number of unexplored points in \( \mathcal{W} \) while accurately mapping obstacles and free regions.

\subsection{Robot Dynamics}
Each robot \( i \) is characterized by its position, motion, and obstacle detection capabilities. At time \( t \), the robot’s position is represented as:
\[\mathbf{p}_i(t) = (x_i(t), y_i(t)) \in \mathcal{W},
\]
where \( x_i(t) \) and \( y_i(t) \) are the robot’s coordinates in the 2D environment. The position \( \mathbf{p}_i(t) \) evolves over time based on the robot’s velocity.

The motion of the robot is governed by its velocity vector \( \mathbf{v}_i(t) \), and the position at the next time step is updated as:
\[
\mathbf{p}_i(t+1) = \mathbf{p}_i(t) + \mathbf{v}_i(t) \Delta t,
\]
where \( \Delta t \) is the time step.


During exploration, robots use tactile sensors to detect obstacles upon physical contact. When an obstacle is detected, the robot logs its current position, creating a local set of obstacle points:
\[
\mathcal{O}_i = \{ \mathbf{p}_i(t) \mid \text{ObstacleFlag} = 1 \}.
\]
This set represents all obstacle points detected by the robot during its operation.



\subsection{Collaborative Mapping}
The robots first work individually to build local point cloud and aggregate them together to build a global map. The global obstacle map is constructed by aggregating local obstacle detections from all robots:
\[
\mathcal{O} = \bigcup_{i=1}^N \mathcal{O}_i,
\]
where \( \mathcal{O} \) represents the global set of obstacle points, and \( N \) is the total number of robots. Exploration is considered complete when the number of local detected obstacle points \( |\mathcal{O}_i| \) reaches a predefined threshold \( M \) such that $
|\mathcal{O}_i| \geq M,
$
where \( M \) is the target number of obstacle points required to sufficiently map the environment.


\subsection{Exploration Objectives}
The exploration task is driven by two key objectives. 


The first objective is to minimize redundancy by reducing the revisits to already explored locations. For each robot \( i \), the path it follows over time is represented as:
\[
\mathcal{P}_i = \{ \mathbf{p}_i(t) \mid t = 0, 1, \dots, T \}.
\]
The redundancy cost for a robot is calculated as:
\[
R_i = \sum_{\mathbf{p}_i \in \mathcal{P}_i} \mathbb{1}\{ \mathbf{p}_i \text{ already visited} \},
\]
where \( \mathbb{1}\{\cdot\} \) is the indicator function that evaluates to 1 if the location has been visited previously, and 0 otherwise.

The second objective is to avoid collisions between robots. To achieve this, each pair of robots must maintain a minimum separation distance \( d_{\text{min}} \) at all times:
\[
\|\mathbf{p}_i(t) - \mathbf{p}_j(t)\| \geq d_{\text{min}}, \quad \forall i \neq j.
\]
This ensures safe and efficient operation of the swarm in the shared workspace.


\subsection{Constraints}
The exploration task is constrained by communication limits, where robots can exchange information only within a specified communication range \(r_{\text{comm}}\). To account for this, the neighbor set \(\mathcal{N}_i\) of each robot \(i\) is defined as:  
\[
\mathcal{N}_i = \{j \mid \|\mathbf{p}_i(t) - \mathbf{p}_j(t)\| \leq r_{\text{comm}}, \quad \forall i, j\}.
\]

\vspace{-2mm}
\subsection{Problem Statement}
\vspace{-2mm}
Given the environment \( \mathcal{W} \subset \mathbb{R}^2 \),
 \( N \) robots with initial positions \( \{ \mathbf{p}_1(0), \mathbf{p}_2(0), \dots, \mathbf{p}_N(0) \} \), find the trajectories \( \{\mathbf{p}_i(t)\}_{t=0}^T \) of all robots such that:
\begin{enumerate}
    \item Redundancy \( R_i \) is minimized for each robot.
    \item Collisions are avoided.
    \item A global map \( \mathcal{M} \) is constructed based on point clouds from each robot.
\end{enumerate}


\section{Methodology}
This section describes the proposed methodology for decentralized multi-robot exploration and mapping. Each robot selects goals, evaluates costs, and navigates autonomously while sharing information to collaboratively construct a global map.
\subsection{Candidate Goal Generation}
At each time step \( t \), each robot \( i \) generates a set of candidate goals \( \mathcal{G}_i = \{\mathbf{g}_1, \mathbf{g}_2, \dots, \mathbf{g}_P\} \) around its current position \( \mathbf{p}_i(t) \). These goals are distributed evenly on a circle of radius \( r \), centered at the robot's current position:
\[
\mathbf{g}_k = \mathbf{p}_i(t) + r \begin{bmatrix}
\cos \theta_k \\ 
\sin \theta_k
\end{bmatrix}, \quad \theta_k = \frac{2\pi k}{P}, \, k = 1, \dots, P,
\]
where \( P \) is the total number of candidate goals, and \( \theta_k \) is the angular separation between consecutive goals.
\subsection{Cost Function}
To evaluate each candidate goal \( \mathbf{g}_k \in \mathcal{G}_i \), a cost function \( J_i(\mathbf{g}_k) \) is defined. The cost function incorporates the following terms:

\paragraph{Collision Cost} The collision cost penalizes goals that are close to neighboring robots:
\[
J_{\text{collision}} = \sum_{j \in \mathcal{N}_i} \frac{1}{\|\mathbf{g}_k - \mathbf{p}_j(t)\|},
\]
where \( \mathbf{p}_j(t) \) is the position of the \( j\)-th neighboring robot.

\paragraph{Redundancy Cost} The redundancy cost penalizes revisiting points that have already been logged as obstacles:
\[
J_{\text{redundancy}} = \sum_{\mathbf{o} \in \mathcal{O}_i} \frac{1}{\|\mathbf{g}_k - \mathbf{o}\|^2},
\]
where \( \mathcal{O}_i \) is the set of obstacle points logged by robot \( i \) during its exploration.

\paragraph{Total Cost} The total cost \( J_i(\mathbf{g}_k) \) for each candidate goal is computed as:
\[
J_i(\mathbf{g}_k) = \beta J_{\text{collision}} + \gamma J_{\text{redundancy}},
\]
where \( \beta \) and \( \gamma \) are weighting factors that determine the relative importance of collision avoidance and redundancy minimization.

\subsection{Algorithm Overview}
The distributed optimization-based algorithm for multi-robot exploration operates iteratively, enabling each robot to determine its next exploration goal while minimizing conflicts and redundancies. The algorithm proceeds in distinct steps, as outlined in Algorithm~\ref{alg:goal_selection}.

\begin{algorithm}

\caption{Distributed Optimization-Based Goal Selection for Multi-Robot Exploration}
\label{alg:goal_selection}
\begin{algorithmic}[1]

\State \textbf{Initialization:}
\State Initialize the positions of all robots: \(\{p_1, p_2, \dots, p_n\}\).
\State Define the set of logged points for each robot: \(O_i = \{o_{1}, o_{2}, \dots, o_{m}\}\).
\State Define communication range \(r_{\text{comm}}\).
\State Set circle parameters: radius \(r\) and number of sampled points \(k\).
\State Generate initial candidate goals for each robot: \(\mathcal{G}_i = \{g_{1}, g_{2}, \dots, g_{k}\}\).
\State Set weight parameters: \(\beta\) (collision avoidance) and \(\gamma\) (redundancy minimization).
\State Set number of points \(p\) to log before termination.

\State \textbf{{Step 1: Compute Neighbor Set of each robot \(i\) based on Distance to other Robots}}
\[
\mathcal{N}_i = \{j \mid \|\mathbf{p}_i(t) - \mathbf{p}_j(t)\| \leq r_{\text{comm}}, \quad \forall i, j\}
\]


\State \textbf{Step 2: Compute Local Cost Function}
\For{each candidate goal \(g \in \mathcal{G}_i\)}
    \State Compute the local cost: 
    \[
    J_i(g_k) = \beta \sum_{j \in \mathcal{N}_i} \frac{1}{\|g_k - p_j\|} 
    + \gamma \sum_{o \in O_i} \frac{1}{\|g_k - o\|}
    \]
\EndFor


\State \textbf{Step 3: Minimize Local Cost Function}
\State Select the candidate goal \(g_i\) that minimizes the local cost:
\[
g_i = \arg\min_{g \in \mathcal{G}_i} J_i(g_k)
\]


\State \textbf{Step 4: Move Toward Selected Goal}
\State Assign the resolved goal \(g_i\) to robot \(i\).
\State Update the robot's position using a motion model:
\[
p_i(t+1) = p_i(t) + v_i(t) \Delta t
\]


\State \textbf{Step 5: Repeat Until Termination Condition}
\While{the robot has logged fewer than \(p\) obstacle points}
    \For{each robot \(i\)}
        \If{robot \(i\) reaches \(g_i\) or encounters an obstacle}
            \If{robot \(i\) collides with another robot \(j\)}
                \State Move back
            \Else
                \State Log the collision point to local storage
            \EndIf
            \State Recompute candidate goals \(\mathcal{G}_i\)
        \EndIf
    \EndFor
\EndWhile
\State Stop robot movement at the current position.
\end{algorithmic}
\end{algorithm}

The algorithm begins by initializing the positions of all robots and defining key parameters, such as the communication range \(r_{\text{comm}}\) and goal sampling settings (\(r\) and \(k\)). Each robot generates a set of candidate goals \(\mathcal{G}_i\) and calculates the local cost function \(J_i(g)\) for each candidate.

Each robot selects the candidate goal that minimizes its local cost function and moves toward it. When an obstacle is encountered, the robot logs the collision point. New candidate goals are computed either when the robot encounters an obstacle or reaches its previously assigned goal. This process continues until each robot logs a predefined number of obstacle points, at which point the robot stops moving.

\section{Experimental Design and Results}
\begin{figure}[h!]
    \centering
    \includegraphics[width=0.9\linewidth]{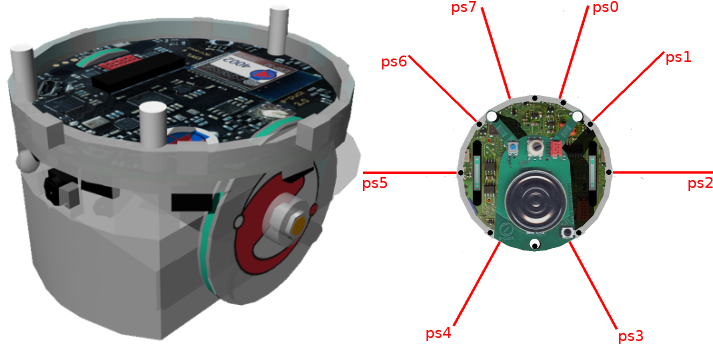}
    \caption{Left: The e-puck robot used in the experiments. Right: Top view of the e-puck showing the distribution of rangefinder sensors (ps0-ps7) used for obstacle detection.}
    \label{Fig:sensor}
\end{figure}

To conduct the experiments, we utilized e-puck robots within the Webots simulation environment \cite{michel2004cyberbotics}. Each robot was equipped with rangefinder sensors positioned around its body to detect obstacles as shown in Figure 2. To emulate the functionality of tactile sensors, the detection range of the rangefinders was reduced, ensuring that obstacles were only identified when the robot was in close proximity, effectively simulating physical contact. The goal of the experiment was to explore and map a 1.5 × 1.5 m environment containing three obstacles.

To evaluate the proposed approach, three sets of experiments were conducted. In all experiments, the radius \(r\) for sampling candidate goals \(\mathcal{G}_i\) was fixed at 0.3 m, while the number of sampled points \(k\) was set to 360. For the first two experiments, the communication radius \(r_{\text{comm}}\) was set to 3 m to ensure the communication graph remained fully connected throughout the simulations. These parameter values were selected to achieve a balance between navigation precision and computational efficiency. Each simulation concluded once a robot logged 100 points, ensuring adequate map coverage within a manageable simulation duration.
\subsection{Parameter Evaluation for Weight Coefficients \(\beta\) and \(\gamma\)}
The first experiment analyzed the effects of varying the weight parameters \(\beta\) and \(\gamma\), which regulate the trade-off between collision avoidance (\(\beta\)) and exploration of unvisited areas (\(\gamma\)). These parameters were tested at three levels (0.1, 0.5, and 0.9) to represent low, medium, and high prioritization of their respective objectives.

\begin{table}[h!]
\caption{Experiments with Different Parameter Values for \(\beta\) and \(\gamma\) in 3 Robots Simulation and Fully Connected Graph}
\centering
\resizebox{0.5\textwidth}{!}{%
\begin{tabular}{|c|c|c|c|c|}
\hline
\textbf{Number of Robots} & \makecell{\textbf{Simulation} \\ \textbf{Time (s)}} & \makecell{\textbf{Number of} \\ \textbf{Robot Collisions}} & \makecell{\textbf{Number of} \\ \textbf{Logged Points}} & \makecell{\textbf{Logged Points} \\ \textbf{per Second}} \\ \hline
\(\beta = 0.1, \gamma = 0.1\) & 248 & 0 & 261 & 1.05\\ \hline
\(\beta = 0.1, \gamma = 0.5\) & 350 & 0 & 258 & 0.74\\ \hline
\(\beta = 0.1, \gamma = 0.9\) & 308 & 65 & 230 & 0.75\\ \hline
\(\beta = 0.5, \gamma = 0.1\) & 300 & 0 & 292 & 0.97\\ \hline
\(\beta = 0.5, \gamma = 0.5\) & 260 & 0 & 261 & 1.00\\ \hline
\(\beta = 0.5, \gamma = 0.9\) & 328 & 19 & 237 & 0.72\\ \hline
\(\beta = 0.9, \gamma = 0.1\) & 240 & 0 & 300 & 1.25\\ \hline
\(\beta = 0.9, \gamma = 0.5\) & 339 & 0 & 251 & 0.74\\ \hline
\(\beta = 0.9, \gamma = 0.9\) & 258 & 0 & 261 & 1.01\\ \hline
\end{tabular}%
}
\label{tab:table_1}
\end{table}

Table~\ref{tab:table_1} presents the results. The optimal parameter combination was identified as \(\beta = 0.9\) and \(\gamma = 0.1\), which achieved 300 logged points without any collisions while minimizing simulation time. This combination demonstrated the best trade-off, as it allowed robots to meet the termination criteria efficiently. Other parameter configurations failed to meet the termination criteria due to robots becoming trapped in local minima. This behavior occurred because once a robot met its termination criteria, it stopped moving, while other robots continued to prioritize maximizing distance from the stationary robot and previously explored areas. This often resulted in repetitive movements (e.g., circling or oscillating along a line) without exploring new regions.

Collision avoidance was generally effective except for scenarios with the largest priority priority on exploration (\(\beta = 0.1, \gamma = 0.9\) and \(\beta = 0.5, \gamma = 0.9\)). These configurations also resulted in the lowest rates of logged points per second (\(0.72\) and \(0.75\), respectively).

\begin{figure}
    \centering
    \includegraphics[width=\linewidth]{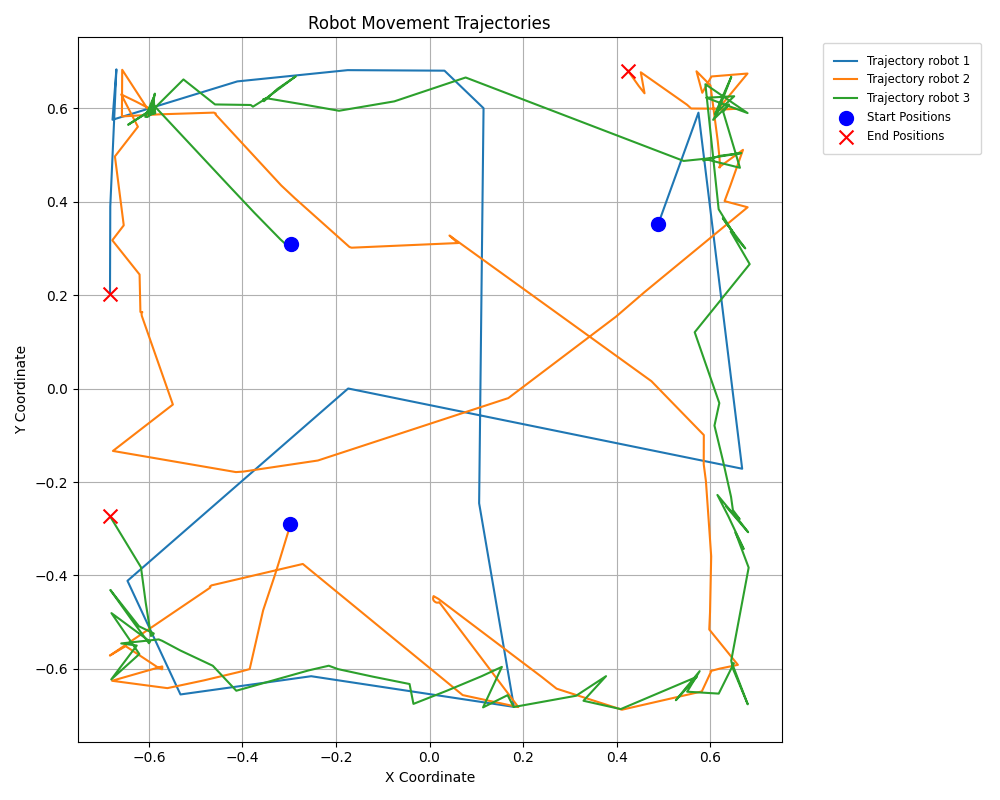}
    \caption{Robot movement in the 3 robot simulation with the best weighting parameters (\(\beta = 0.9\) and \(\gamma = 0.1\)). It can be observed that the robots tend to stay at the borders of the environment, based on the high collision avoidance priority given.}\label{Fig:RobotMovement}
\end{figure}

Figure \ref{Fig:RobotMovement} presents the robot trajectories for the optimal parameter configuration (\(\beta = 0.9\), \(\gamma = 0.1\)). The trajectories reveal a tendency for robots to navigate predominantly along the environment's periphery, resulting in reduced exploration of central regions where obstacles are located.

This behavior highlights a trade-off inherent in the chosen parameter configuration. The high priority placed on collision avoidance (\(\beta = 0.9\)) encourages robots to maintain greater distances from each other, reducing the likelihood of collisions. However, this also leads to less frequent traversal of central areas, resulting in a lower resolution of obstacle mapping in these regions. In contrast, parameter configurations that prioritize exploration (\(\gamma > \beta\)) facilitate more uniform coverage of the environment but at the expense of increased collision risk. This underscores the importance of carefully balancing collision avoidance and exploration objectives to achieve comprehensive and efficient mapping.
\subsection{Evaluation with Varying Numbers of Robots}
In the second experiment, the scalability of the proposed approach was assessed by varying the number of robots in the environment. Using the optimal parameters (\(\beta = 0.9, \gamma = 0.1\)) identified in the first experiment, simulations were conducted with 2, 3, 5, 7, and 10 robots.

\begin{table}[h!]
\caption{Experiments with Different Numbers of Robots and Fully Connected Graph (\(\beta = 0.9\) and \(\gamma = 0.1\))}
\centering
\resizebox{0.5\textwidth}{!}{%
\begin{tabular}{|c|c|c|c|c|}
\hline
\textbf{Number of Robots} & \makecell{\textbf{Simulation} \\ \textbf{Time (s)}} & \makecell{\textbf{Number of} \\ \textbf{Robot Collisions}} & \makecell{\textbf{Number of} \\ \textbf{Logged Points}} & \makecell{\textbf{Logged Points} \\ \textbf{per Second}} \\ \hline
2 & 199 & 0 & 165 & 0.83 \\ \hline
3 & 240 & 0 & 300 & 1.25 \\ \hline
5 & 443 & 0 & 500 & 1.13 \\ \hline
7 & 433 & 0 & 700 & 1.62 \\ \hline
10 & 560 & 0 & 1000 & 1.44 \\ \hline
\end{tabular}%
}
\label{tab:table_2}
\end{table}

The results, summarized in Table~\ref{tab:table_2}, indicate that increasing the number of robots improves the logged points per second. However, the optimal number of robots for this specific environment was determined to be 7. While additional robots increased logged points, they also introduced challenges such as computational overhead and potential goal selection conflicts. In the case of two robots, one robot completed its logging goal quickly, leaving the other robot in a local minimum, unable to explore new areas effectively.

\begin{figure}

    \centering
    \includegraphics[width=\linewidth]{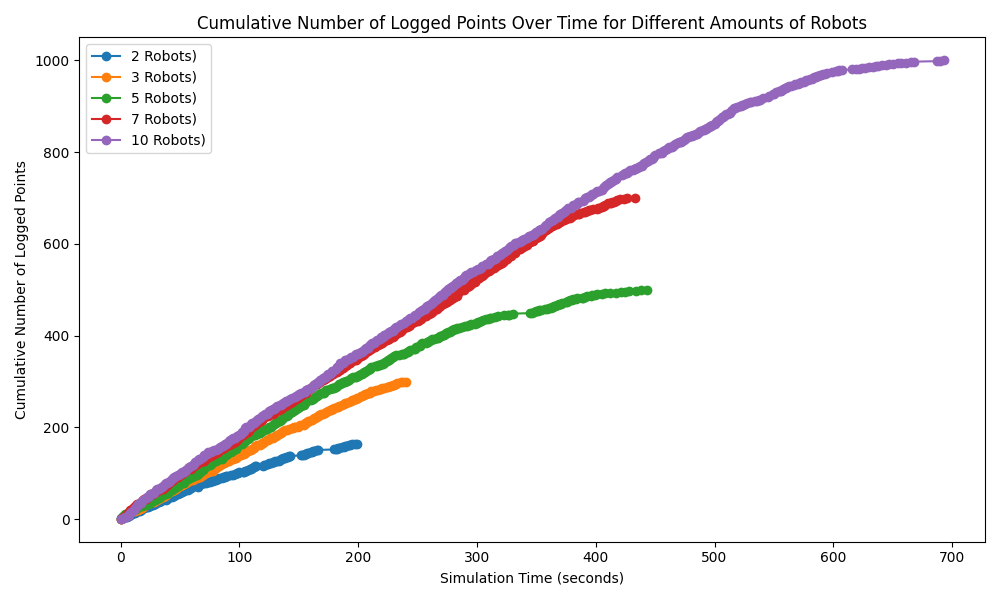}
    \caption{Logged points over time for simulations with different amounts of robots.}\label{Fig:LoggedPointsOverTime}\vspace{-5mm}
\end{figure}

\begin{figure}

    \centering
    \includegraphics[width=\linewidth]{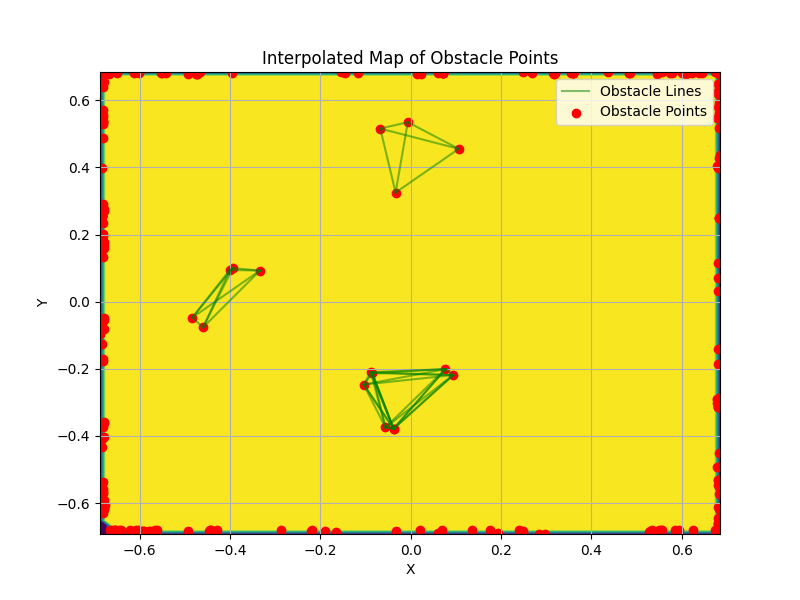}
    \caption{Interpolated map from the logged obstacle points from the 10 robots simulation with \(\beta = 0.9\) and \(\gamma = 0.1\).}\label{Fig:InterpolatedMap}
\end{figure}

Figure~\ref{Fig:LoggedPointsOverTime} illustrates the progression of logged points over time for varying team sizes. The results demonstrate an approximately linear increase in logged points initially, which gradually slows as the maximum point count is approached. Among the configurations, the simulation with seven robots exhibited the fastest convergence to the maximum logged points, underscoring the efficiency of this team size for the given environment.

Interestingly, the configuration with five robots deviated from the observed trend, reaching its maximum logged points at a slower rate compared to the three-robot setup. This suggests the presence of diminishing returns in performance efficiency as the team size increases beyond a certain threshold. Such findings indicate that while increasing the number of robots can enhance coverage and total logged points, it may also introduce inefficiencies related to coordination and resource allocation.

Figure~\ref{Fig:InterpolatedMap} presents the interpolated map generated from the logged obstacle points in the 10-robot simulation. As expected, the higher number of robots resulted in the highest total number of logged points. However, the resolution of inner objects was not significantly improved compared to some of the three-robot configurations. This outcome can be attributed to the high priority placed on collision avoidance, which drove the robots toward the boundaries of the environment, limiting their exploration of inner regions.

Interestingly, the inner environment was explored more frequently in the 10-robot simulation than in the configurations with five or seven robots. This increased exploration is likely due to the high density of robots within the limited environment space. The greater number of agents increased the likelihood of overcoming the collision-avoidance priority, as robots with limited exploration opportunities near the boundaries were more inclined to turn inward. In contrast, simulations with fewer robots tended to focus more on the border areas, resulting in sparser coverage of the inner regions.
\subsection{Communication Evaluation with Distance Constraints}
The third experiment examined the impact of varying communication ranges on robot coordination. Communication was defined by a fixed distance parameter \(r_{\text{comm}}\), representing the maximum range within which robots could exchange relative positional information. Robots beyond this range were unable to communicate.

This evaluation used the seven-robot configuration, previously identified as the most efficient in terms of exploration rate for this specific environment. The tested communication distances were 0.4 m, 0.5 m, 0.6 m, 0.7 m, and 0.8 m. These values exceeded the sampling radius for candidate goals, ensuring periodic communication disruptions even at the largest distance, thereby simulating realistic constraints.

\begin{table}[h!]
\caption{Experiments with Different Maximum Communication Distances for 7-Robot Simulation (\(\beta = 0.9\) and \(\gamma = 0.1\))}
\centering
\resizebox{0.5\textwidth}{!}{%
\begin{tabular}{|c|c|c|c|c|}
\hline
\makecell{\textbf{Maximum Communication} \\ \textbf{Distance (m)}} & \makecell{\textbf{Simulation} \\ \textbf{Time (s)}} & \makecell{\textbf{Number of} \\ \textbf{Robot Collisions}} & \makecell{\textbf{Number of} \\ \textbf{Logged Points}} & \makecell{\textbf{Logged Points} \\ \textbf{per Second}} \\ \hline
0.4 & 886 & 35 & 700 & 0.79 \\ \hline
0.5 & 894 & 17 & 700 & 0.79 \\ \hline
0.6 & 720 & 11 & 700 & 0.97 \\ \hline
0.7 & 735 & 5 & 700 & 0.95 \\ \hline
0.8 & 580 & 0 & 700 & 1.21 \\ \hline
\end{tabular}%
}
\label{tab:table_3}\vspace{-3mm}
\end{table}

The results in Table~\ref{tab:table_3} highlight key trends. As the communication distance decreases, the number of robot collisions increases. This outcome is expected, as shorter communication ranges limit the time available for robots to react to neighboring movements, thereby raising the collision likelihood.

\begin{figure}
    \centering
    \includegraphics[width=\linewidth]{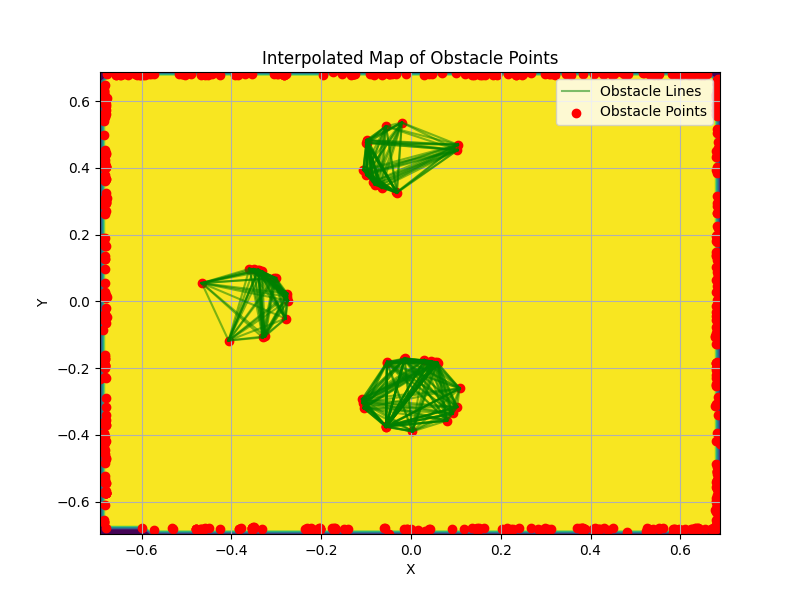}
    \caption{Interpolated map from the logged obstacle points in the 7-robot simulation with communication constraints (\(r_{\text{comm}} = 0.5~\text{m}\)).}\label{Fig:InterpolatedMap7_best}
\end{figure}

Furthermore, smaller communication distances result in longer exploration times. However, these configurations often yield more detailed coverage of the environment, maximizing overall mapping resolution. This trade-off is illustrated in Figure~\ref{Fig:InterpolatedMap7_best}, where the interpolated map from the 7-robot simulation with \(r_{\text{comm}} = 0.5~\text{m}\) demonstrates higher density compared to Figure~\ref{Fig:InterpolatedMap}, which represents a fully connected communication graph at all times with 10 robots.

\subsection{Challenges}
The exploration and mapping process posed several challenges, including accurate emulation of tactile sensing using rangefinder sensors and ensuring seamless multi-robot coordination. The reduced range of sensors, designed to mimic physical contact, requires careful calibration to avoid missed detections or false positives, particularly in dynamic and cluttered environments. Communication among robots, essential for distributed optimization and global map construction, was another critical challenge, as delays, packet loss, or range limitations could impact the quality and timeliness of shared data. Ensuring collision-free navigation while minimizing redundancy in goal assignments was further complicated by the dynamic interplay of multiple robots operating in close proximity. Additionally, the computational load of cost evaluations and global cost aggregation increased with the number of candidate goals and robots, presenting scalability issues. Since our implementation was on simulation, we did not consider it much. However, it would be noteworthy to consider if real robots were deployed. 

\section{Conclusion and Future Work}
This study demonstrated the effectiveness of a decentralized multi-robot system in exploring and mapping unknown environments using bioinspired tactile sensing. By utilizing e-puck robots equipped with rangefinder sensors to emulate tactile capabilities, the system successfully navigated a 1.5 × 1.5 m environment containing three obstacles, collaboratively constructing a 2D map. The experiments revealed that optimal parameter tuning for collision avoidance ($\beta$) and redundancy minimization ($\gamma$) significantly impacts the system's performance, with the configuration $\beta$ = 0.9 and $\gamma$ = 0.1 yielding the best results. The approach effectively minimized collisions, reduced redundant movements, and achieved high coverage, validating the system's scalability and robustness in dynamic multi-robot environments.

Future research will focus on enhancing the scalability and adaptability of the proposed system. Expanding experiments to larger, more complex environments with varying obstacle densities will provide deeper insights into the system's capabilities. Real-world deployment with advanced tactile sensors can address limitations of simulated rangefinder-based detection. Further, optimizing to reduce latency and investigating energy-efficient navigation strategies will enhance operational reliability. Integrating dynamic obstacle handling and exploring machine learning techniques for adaptive decision-making are promising directions to improve system performance in diverse real-world applications, such as search-and-rescue and industrial inspection.
\section{Project Contribution}
Roman: tuning of rangefinder sensors to emulate tactile sensing; development of goal selection mechanism and local navigation strategies; ideation of the cost function with parameters; debugging and performance analysis of the exploration system; writing the report.

Jannik: setup and integration of e-puck robots in Webots simulation environment; implementation of communication framework for distributed optimization; algorithm design and implementation; refinement of system behavior based on experimental results; validation and analysis of cost function parameters; writing the report.

\bibliographystyle{IEEEtran}
\bibliography{references.bib}

\end{document}